*A Pre-print*

# Investigating Wave Energy Potential in Southern Coasts of the Caspian Sea Using Grey Wolf Optimizer Algorithm

Erfan Amini[1], Seyed Taghi Omid Naeeni[1], Pedram Ghaderi[2], Fereidoun Amini[2*]

[1] Coastal and Offshore Structures Engineering Group, College of Engineering, University of Tehran, Tehran, Iran.
[2] Department of Civil Engineering, Iran University of Science and Technology, Tehran, Iran.

*Corresponding author
*famini@iust.ac.ir*

**Abstract:** There is a significantly accelerating trend in the application of the marine wave energy converters in recent years. As a result, it is imperative to adopt a suitable point for implementing these systems. Besides, the Caspian Sea, as one of the most important marine renewable energy sources in Asia, is capable of supplying the coastal areas with a large amount of energy. Therefore, areas around nine ports in the southern coasts of the Caspian Sea were selected to measure their wave energy potential. Initially, the amount of energy on these points was measured using the irregular energy theory. It was observed that the wave power was higher in the southwestern areas (within the Kiashahr coast and Anzali port) than the southeastern areas. A new approach was developed to compare these points and measure their fitnesses in supplying the maximum energy using the Grey Wolf optimizer (GWO) algorithm and time history analysis. In this method, the optimal parameters were first extracted from the algorithm for assessing the points within the southern areas of the Caspian Sea. These values were regarded as the assessment indices. Then, the fitness of each point was obtained using the correlation function and the norm vector to present the most optimal position with maximum wave energy exploitation potential. This new approach was validated with analytical data, and its accuracy in predicting and comparing the wave power on different points was approved. Finally, by a side-by-side comparison of the parameters affecting the wave energy, the optimum range of significant wave height and wave energy period was achieved.

# 1 Introduction

Nowadays, with the progress of technology and increasing need of human societies for energy as well as reduced hydrocarbon resources and fossil fuels, we have observed exploitation of natural resources such as seas and oceans for energy supply[1]. Non-tidal waves in the sea with periods between 0.5 and 30 seconds have a high potential for energy transport and supply[2]. Deep-water waves travel long distances without energy loss to reach shallower areas that are accessible to humans. The energy capacity of seas and oceans has been estimated at 2.7 TW. By examining global wind-wave model data, Mork et al. [3] presented the results of wave energy potential distribution in seas and oceans as well as the seasonality of gross theoretical wave power distribution, as shown in Fig.1.

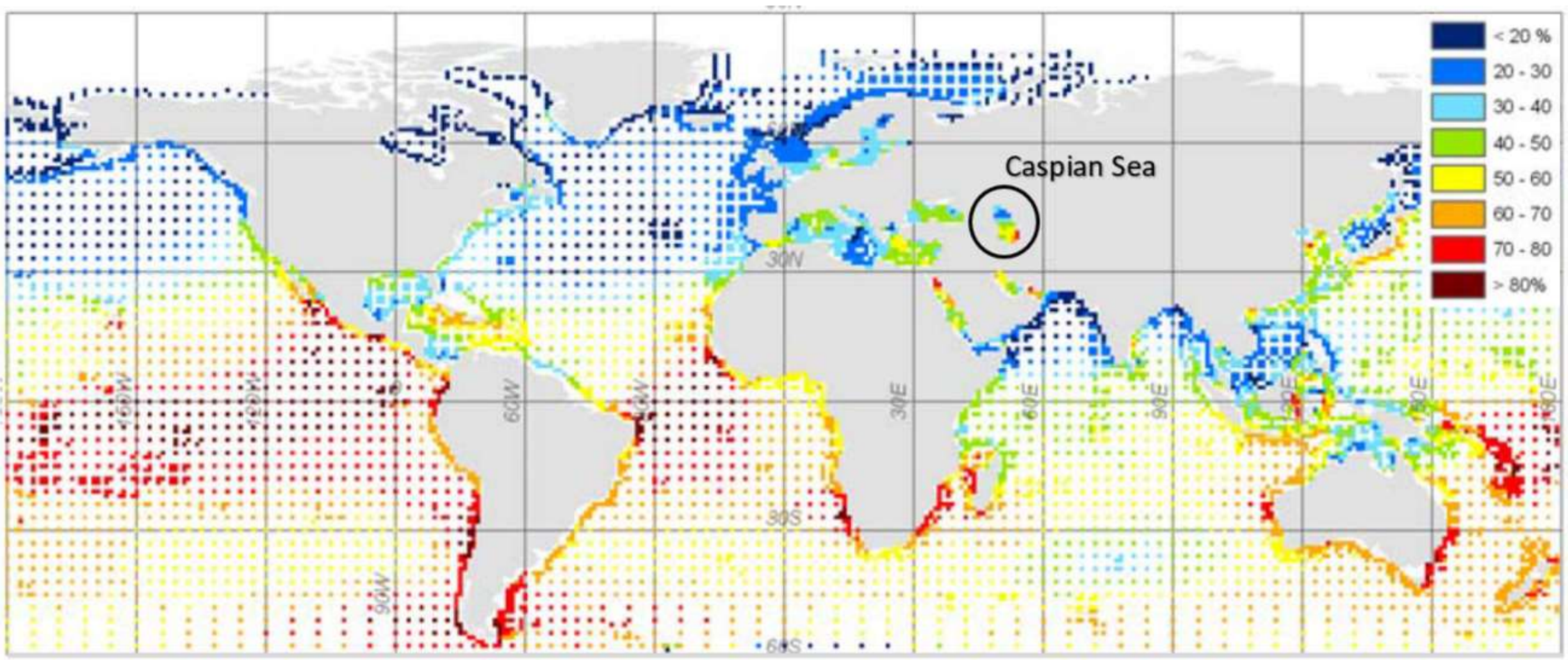

**Figure 1:** Seasonality of gross theoretical wave power distribution [3] (ratio of minimum monthly wave power and annual wave power); Low percentage shows high seasonality. Areas colored brown, red and orange have relatively stable wave climate regimes.

The Caspian Sea is a body of water bounded by Iran to the south, Russia to the north, Russia and Azerbaijan to the west, and Turkmenistan and Kazakhstan to the east. This sea which is sometimes classified either as the largest lake in the world, or the smallest sea on the Earth, is the largest landlocked body of water. Its length is about 1030 -1200 (km) and its width are from 196 to 435 (km). The Caspian Sea surface is about 28 meters below the free sea level. The northern part of the sea is very shallow so that only half percent of the seawater is located in the northern quarter of the sea, with the average depth of fewer than 5 meters [4]. Therefore, it is more important to investigate its southern shores for wave extraction. Fig. 2 shows the Caspian Sea overview and depth changes prevailing.

The Caspian Sea has always been the focus of the industry as one of the most important energy sources in Asia. The expansion of its southern coast also leads to a high potential for wave energy utilization.

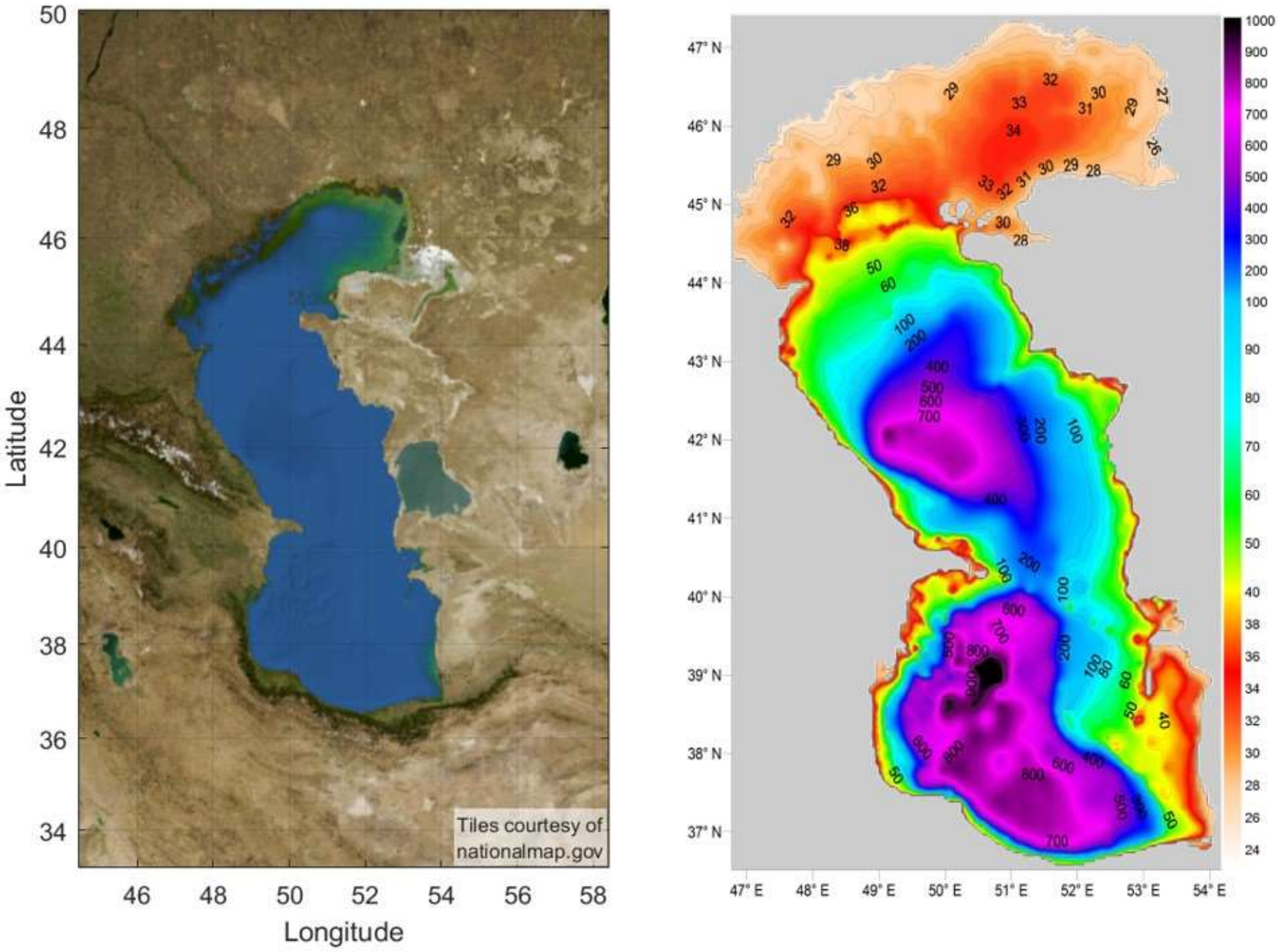

**Figure 2:** The landscape of the Caspian Sea (left), the depth of the Caspian Sea (right) [4]

In Fig. 1, we observe that in the southern part of the Caspian Sea, the wave regime is almost stable during different seasons. It leads to the high possibility of designing different wave energy converters that can constantly work over a year. In the last decade, a variety of studies have been conducted on the potential of wave energy in this sea, although studies on the southern coasts of the sea have been rarely found. Taebi and Golshani [5] used Iranian Seas Wave Modeling (ISWM) to present wave hind-cast modeling for 33 years (from 1975 to 2007) and verify the outputs of their model by satellite and buoy measurements in Iranian Coast. The outputs of this modeling, entitled Iranian Wave Atlas (IWA), provided reliable data for studying the Caspian Sea waves. Abbaspour and Rahimi later supplemented these data by examining the potential of offshore renewable energies in 2011[6]. We also use this information in the present study. Chegini et al. [7] investigated 18 different sites in the southern coasts of the Caspian Sea, Persian Gulf and the Oman Sea to determine and assess the wave energy distribution as well as the average and maximum annual power. Results of this study included the sites of Bandar Anzali, Nowshahr and Amirabad in the Caspian Sea, and Bushehr, Asaluyeh, Jask and Chabahar in the Persian Gulf and Oman Sea. In this work, the maximum annual wave power was in Asaluyeh site in the Persian Gulf and the ports of Nowshahr, Anzali and Amirabad in the southern coasts of the Caspian Sea, respectively. Zabihian and Fung [8] used wind data processing from 1986 to 1995 for 14 sites that were collected on the Caspian and Oman coasts to obtain the wave characteristics. According to studies, the Chabahar coast has the highest potential for wave energy. The site is located on the coast of Oman Sea, which is connected to the Indian Ocean, where the average wave energy is between 10 and 15 kW/m. The islands in the Persian Gulf have a great wave energy potential and, given that these islands are not connected to Iran's electricity grid, the energy production through the energy of tidal waves in these islands can have economic efficiency[9]. Studies have also been conducted on wave energy for the construction of certain types of wave energy converters. For example, Soleymani and Ketabdari [10] studied the sites of Amirabad in the Caspian Sea as well as Asaluyeh Port and Farur island in the Persian Gulf to absorb the wave energy. Studies have shown that submerged floating elements are very compatible with the conditions of these locations. Moreover, for Farur site, an oscillating water column wave energy converter was very appropriate for the coastline and wave climate. They investigated the potential of power generation of two types of the bottom-fixed heave-buoy array and bottom-fixed oscillating flap wave energy converters for three sites of Amirabad, Farur, and Asaluyeh. The results showed that the best location for three above systems was Farur site because the power output for these three systems was estimated to be 16.8, 96 and 138 kW, respectively. Majidi Nezhad et al. [11] used the annual wave information at six stations in the Caspian Sea in three-hour time steps to estimate the wave energy potential. They calculated the values of significant weight height ($H_s$) and peak wave energy period ($T_p$) in the wavelength units. The results also showed that Bandar Anzali in the southern coasts of the Caspian Sea had a high potential for deploying a wave energy converter system and the energy flux per wavelength unit reached 500-600 W/m.

To exploit energy from the seas, the use of wave energy converters has been the focus of researchers as a reasonable solution for over decades. The optimal performance of wave energy converters and, consequently, the estimation of maximum efficiency has been known as one of the main challenges. Determining the optimum position of the converter operation is considered as the main parameter in solving this problem.

Based on the direction of the dominant sea waves, different analyses have been carried out on the forecasting of waves in the southern areas of the Caspian Sea. Fig. 3 shows the predicted values of 50-year dominant waves in the southern regions of the sea-based on the Gumble distribution. Given the height of the waves shown in Fig. 3, it is essential to use the energy of the waves by finding a point with maximum wave energy and also to have easy access to the beach. Therefore, it is necessary to do comprehensive research that examines the wave data in the southern Caspian Sea to find this point and also provide a general criterion for easier comparison of points' energy using period, wave height and depth. The literature review has shown a few works have been done on the locations close to the Caspian coastline. However, the main issue can be summed up in finding a general criterion to simplify the detection of point fit in providing the maximum amount of energy needed. A new explication based on optimization algorithms is used to solve this issue. In the following, this method is proposed for this problem by combining the regular wave theory with the Grey Wolf Optimizer algorithm. In this method, we find optimal values for the parameters governing the waves (i.e. period and height of the waves and depth) and introduce them as the criteria for evaluating different locations in a range. Finally, the results are compared with those of solving the irregular wave theory equations and verified.

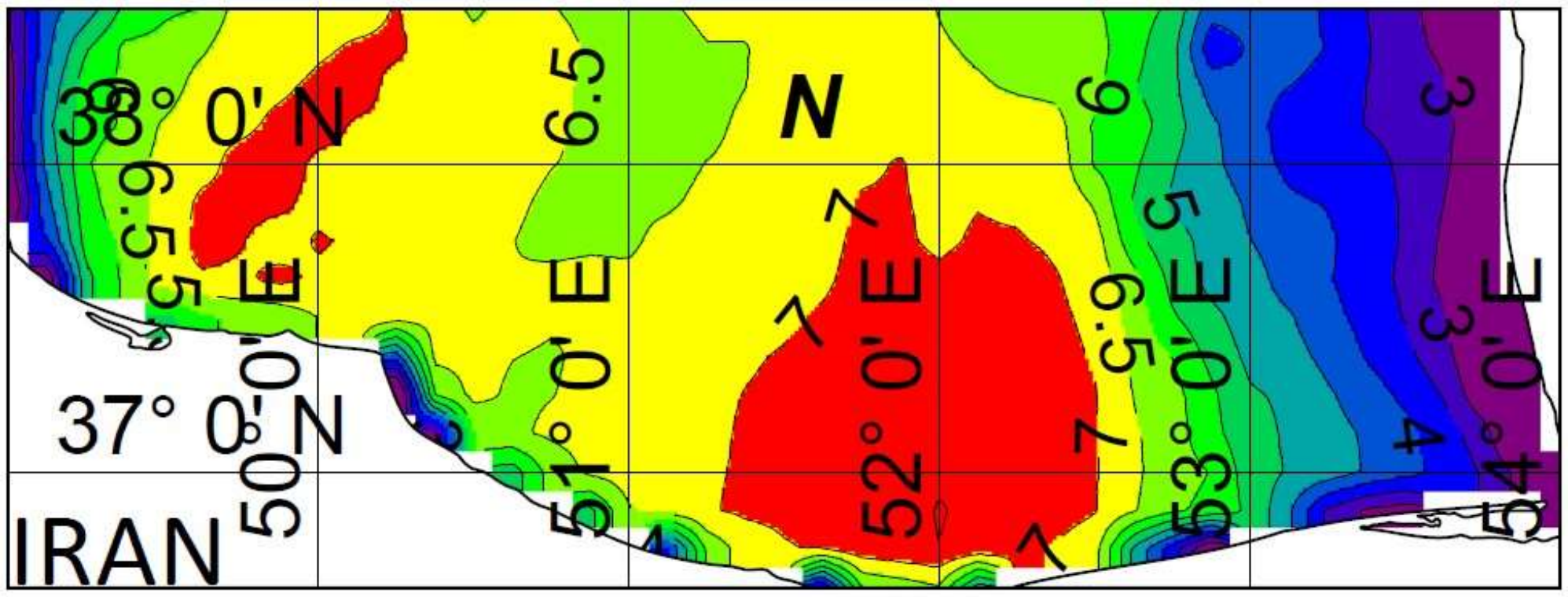

**Figure 3:** 50-year design wave height in prevailing directions in the southern area of the Caspian Sea [5]

## 2 Selection of optimization algorithm

Nowadays, with the progress in wave converters technology, the application of optimization algorithms has led to significant growth in various real engineering problems and, in practice, we observe acceptable results from the outputs of these optimization algorithms in different branches of the wave engineering. For an optimization problem, there might be various solutions; thus, a function called the objective function (simulator or mathematical function) is defined for their comparison and selecting the optimal configurations. Selecting this function depends on the wave-energy problem. Each optimization problem has several independent variables known as design variables. The goal of optimization is to determine the optimal design variables so that the objective function is minimized or maximized [12].

Application of Evolutionary Algorithms (EAs) in the problems related to sea wave energy and wave energy converters has been developed in recent years. In terms of position optimization of WECs, there are considerable achievements which present evolutionary algorithms [13–16] and swarm intelligence methods [17–19] can be sufficiently fast and effective approaches in this way. One of the most important studies in optimizing the WECs is tuning the power take-off settings (PTOs) per each converter, which depends on the WECs location, geometrical parameters and sea states. The application of hybrid methods for optimizing the PTO parameters (spring stiffness and damping coefficient) including a broad set of numerical optimization techniques[20–22], local search [23], evolutionary algorithms [24], surrogate methods [25,26] and cooperative techniques [27] has been increased in the last years as an alternative approach for applying the traditional control methods [28].

However, the grey wolf algorithm has been less used in solving multivariate problems, and the GA and PSO algorithms have been more popular. For instance, Capillo et al. [29] investigated the energy received by a point absorber type inertial sea wave energy converter (ISWEC). Then, using the three optimization scenarios, the dynamic parameters of the converter motion were determined in a way to maximize the amount of energy received by the converter. These three scenarios were GA, PSO algorithm and simultaneous use of these two algorithms. In another research [30], the genetic algorithm and glow worm swarm optimization (GSO) algorithm were employed to verify the proposed method for locating converter construction in a wave energy farm.

The applied optimization algorithms for maximizing the total power output of a wave farm are divided into two categories of heuristic and meta-heuristic. The heuristic algorithms are being caught at local optimum points and early convergence to these points. Meta-heuristic algorithms are presented to solve these problems of heuristic algorithms. In fact, meta-heuristic algorithms are from local optimum points and, along with simplicity and flexibility in solution, we can apply them in a wide range of wave energy problems [31]. One of the criteria for the categorization of optimization algorithms is the number of solutions evaluated in each cycle of algorithm iteration. Accordingly, the algorithms are divided into two categories of methods based on a population of solutions referred to as "population-based" algorithms. Grey Wolf Optimization (GWO) algorithm [32] is a recent population-based meta-heuristic algorithm. Grey Wolf Optimization (GWO) algorithm is patterned based on the grey wolf's organization for hunting in nature. A modified algorithm called group grey wolf optimizer (GGWO) has been used in energy convertion recently[33] . This new algorithm can perform better [32] than such famous evolutionary algorithms and swarm intelligence methods like differential evolution [34], PSO [35], firefly algorithm [36], swallow swarm algorithm [37] and fish swarm optimization [38].

Recently, Islam et al. [39] investigated and compared the performance of seven optimization algorithms, including genetic algorithm (GA), PSO, glowworm swarm optimization (GSO) [17], artificial bee colony (ABC) [40], firefly algorithm (FFA) [36], cuckoo search optimization (CSO) [41] and GWO, on 22 standard benchmark functions. In the following, the accuracy of each algorithm in providing the closest optimal solution to the existing analytical solution for benchmark functions has evaluated. The accuracy percentage of GA, PSO, ABC, FFA, CSO, GSO and GWO algorithm has been seen at 9%, 59%, 41%, 36%, 36%, 45%, 77% respectively. Thus the GWO performs better than other applied algorithms on average. On the other hand, the GWO algorithm not only provides the global optimum with proper accuracy in the above benchmarks but also can perform considerably at solving real engineering problems [27,42,43]. Thus, in this study, given the superior performance and accuracy of the GWO algorithm, it was attempted to use this algorithm in calculating and comparing the energy of different points in the studied areas.

# 3 Research Methodology

We performed numerical modeling in this study using the analytical model and the GWO algorithm. In each section, the equations were formulated and coded in MATLAB. Finally, the solutions received from the proposed optimization approach were verified.

### 3.1 Step 1: Collection of data from the examined area

To examine the southern coasts of the Caspian Sea, first, the applied data of the Iranian Seas Wave Modeling (ISWM) and the Iranian Wave Atlas (IWA) models which are received from the Iranian National Institute for Oceanography and Atmospheric Science (INIO) are analyzed. These datasets belonged to the entire Caspian Sea in a five-year interval (from January 2006 to December 2010) with 1-hour time steps. Therefore, nine ports were selected based on the available technical literature and according to the natural local assessment in southern coasts of the Caspian Sea. Around these ports, the area with the longitude and latitude of 0.2 was assumed, and we extracted the locations with data inside this area. In total, 105 points with data in the southern coasts area of the Caspian Sea were extracted, with specifications given in Fig.4 and Fig.5 The longitude and latitude of these points are presented in Table 1.

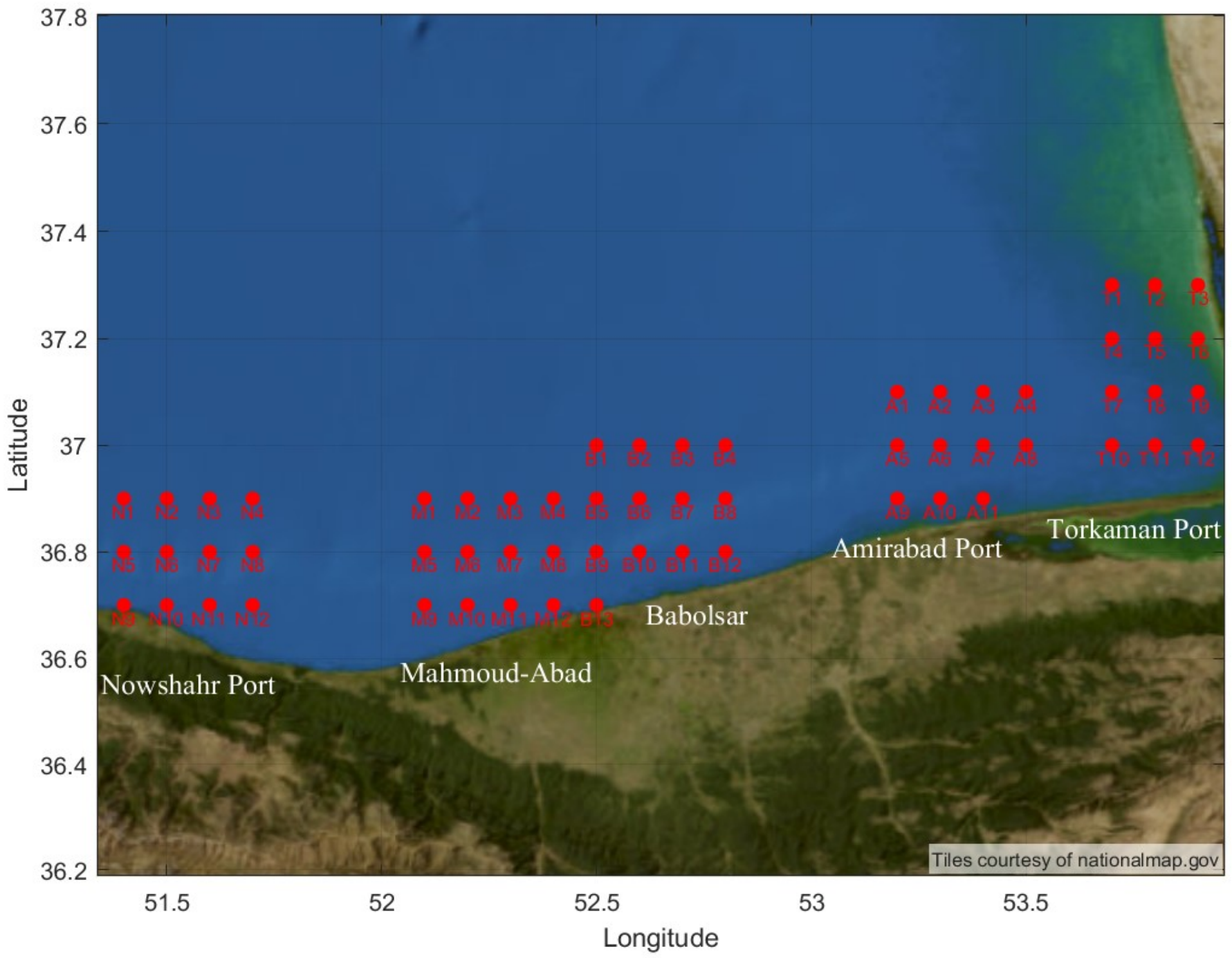

**Figure 4:** A schema of the examined locations in the southeastern coasts of the Caspian Sea (Torkaman, Amirabad, Babolsar, Mahmoud-Abad, and Nowshahr Port).

### 3.2 Step2: Analytical approach based on irregular wave theory

In this case, the approach is developed based on water level fluctuations and wave spectral analysis. Since in the real sea states, the changes in wave height have not been regularized and have shown nonlinear and irregular changes over time, this method has been used for more accurate estimation and more realistic. In this method, first, a complete-time history of the locations is considered. Next, by characterizing the wave spectrum at each point, we study the properties of the waves through spectral analysis. Finally, the amount of wave power at each grid point is extracted, and the parameters obtained are entered into the GWO optimization algorithm to perform the optimization process. We applied the following procedure in the analysis of waves based on the irregular theory. We consider a wave record with duration ($D$) We can accurately reproduce that record as the sum of a large (theoretically infinite) number of harmonic wave components (a Fourier series) as:

$$\xi(t) = \sum_{i=1}^{N} a_i \, cos(2\pi f_i t + a_i) \qquad (f_i = i/D) \tag{1}$$

With a Fourier analysis, we can determine the amplitudes and phases $a_i$ for each frequency. For wave records, the phases have any value between 0 and $2\pi$ with no preference for none value.

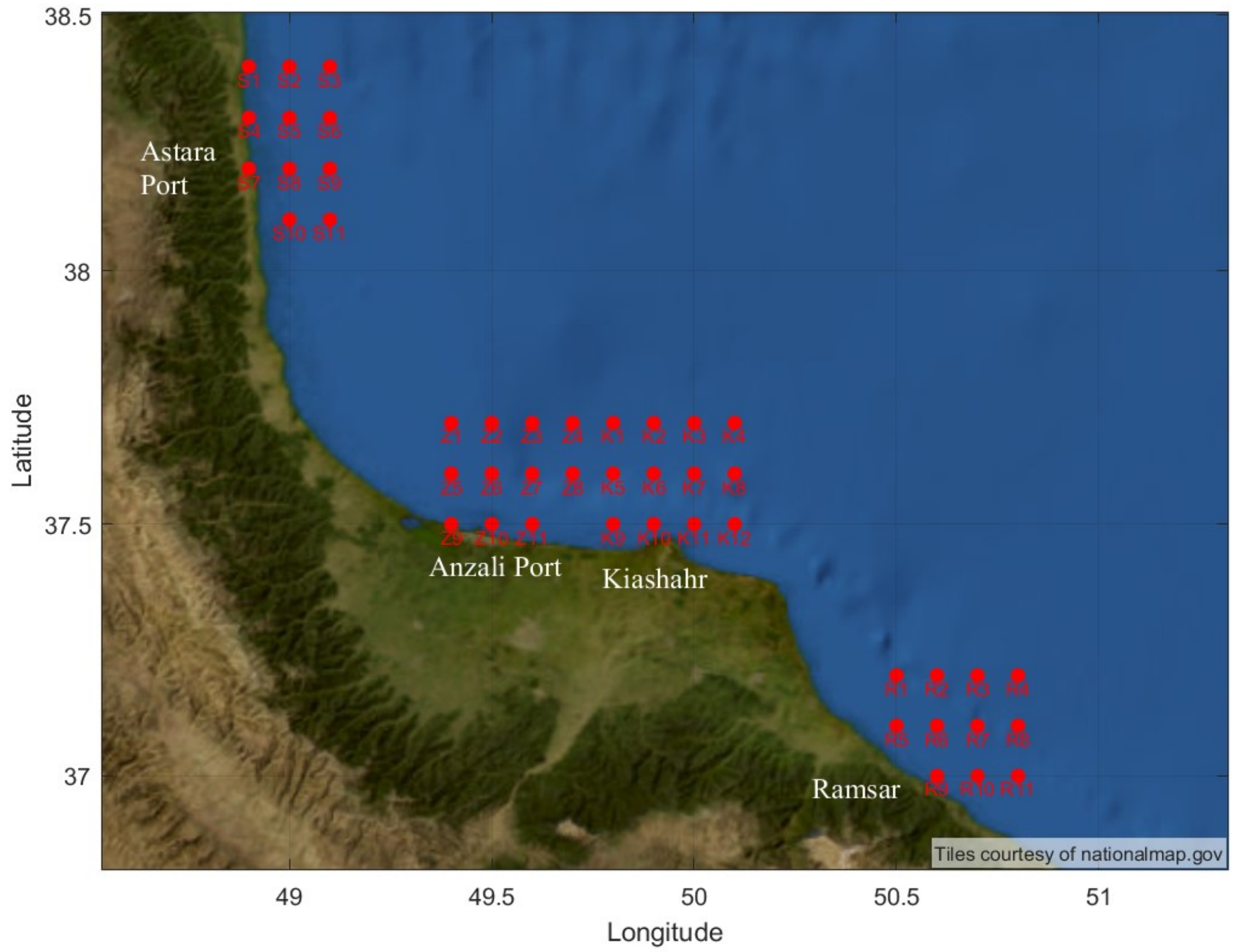

**Figure 5:** A schema of the examined locations in the southwestern coasts of the Caspian Sea (Ramsar, Kiashahr, Anzali Port, and Astara Port).

Only the amplitude spectrum remains to characterize the wave record. To remove the sample character of the spectrum, we should repeat the experiment many times ($M$) and take the average over all these experiments, to find the average amplitude spectrum.

$$\bar{a}_i = \frac{1}{M} \sum_{m=1}^{M} a_{i,m} \qquad (for\ all\ frequencies\ \ f_i) \tag{2}$$

However, it is more meaningful to use the variance of each wave component. An important reason is that the wave energy is proportional to the square of the wave amplitude.

The variance spectrum is discrete, i.e., only the frequencies are presented, whereas all frequencies are present at sea. This is resolved by letting the frequency interval $4f = 1/D \rightarrow 0$. The variance density spectrum is defined as [44]:

$$S_f(f) = \lim_{\Delta f \rightarrow 0} \frac{1}{\Delta f} \frac{1}{2} \overline{a_i^2}, \ \left((m^2.s) \ or \ (m^2/Hz)\right) \tag{3}$$

The variance density spectrum gives a complete description of the surface elevation of ocean waves in a statistical sense, provided that we can see the surface elevation as a stationary Gaussian process. Through this approach, a wave record needs to be divided into segments that are each assumed to be approximately stationary.

**Table 1:** Longitude and latitude of the evaluated points in the Caspian Sea

| | | | | | | | | | | | | | | |
|---|---|---|---|---|---|---|---|---|---|---|---|---|---|---|
| Torkaman port | *Point No.* | 1 | 2 | 3 | 4 | 5 | 6 | 7 | 8 | 9 | 10 | 11 | 12 | - |
| | *Name* | T1 | T2 | T3 | T4 | T5 | T6 | T7 | T8 | T9 | T10 | T11 | T12 | - |
| | *Latitude* | 37.3 | 37.3 | 37.3 | 37.2 | 37.2 | 37.2 | 37.1 | 37.1 | 37.1 | 37 | 37 | 37 | - |
| | *Longitude* | 53.7 | 53.8 | 53.9 | 53.7 | 53.8 | 53.9 | 53.7 | 53.8 | 53.9 | 53.7 | 53.8 | 53.9 | - |
| Amirabad port | *Point No.* | 13 | 14 | 15 | 16 | 17 | 18 | 19 | 20 | 21 | 22 | 23 | - | - |
| | *Name* | A1 | A2 | A3 | A4 | A5 | A6 | A7 | A8 | A9 | A10 | A11 | - | - |
| | *Latitude* | 37.1 | 37.1 | 37.1 | 37.1 | 37 | 37 | 37 | 37 | 36.9 | 36.9 | 36.9 | - | - |
| | *Longitude* | 53.2 | 53.3 | 53.4 | 53.5 | 53.2 | 53.3 | 53.4 | 53.5 | 53.2 | 53.3 | 53.4 | - | - |
| Babolsar | *Point No.* | 24 | 25 | 26 | 27 | 28 | 29 | 30 | 31 | 32 | 33 | 34 | 35 | 36 |
| | *Name* | B1 | B2 | B3 | B4 | B5 | B6 | B7 | B8 | B9 | B10 | B11 | B12 | B13 |
| | *Latitude* | 37 | 37 | 37 | 37 | 36.9 | 36.9 | 36.9 | 36.9 | 36.8 | 36.8 | 36.8 | 36.8 | 36.7 |
| | *Longitude* | 52.5 | 52.6 | 52.7 | 52.8 | 52.5 | 52.6 | 52.7 | 52.8 | 52.5 | 52.6 | 52.7 | 52.8 | 52.5 |
| Mahmoud-Abad | *Point No.* | 37 | 38 | 39 | 40 | 41 | 42 | 43 | 44 | 45 | 46 | 47 | 48 | - |
| | *Name* | M1 | M2 | M3 | M4 | M5 | M6 | M7 | M8 | M9 | M10 | M11 | M12 | - |
| | *Latitude* | 36.9 | 36.9 | 36.9 | 36.9 | 36.8 | 36.8 | 36.8 | 36.8 | 36.7 | 36.7 | 36.7 | 36.7 | - |
| | *Longitude* | 52.1 | 52.2 | 52.3 | 52.4 | 52.1 | 52.2 | 52.3 | 52.4 | 52.1 | 52.2 | 52.3 | 52.4 | - |
| Nowshahr port | *Point No.* | 49 | 50 | 51 | 52 | 53 | 54 | 55 | 56 | 57 | 58 | 59 | 60 | - |
| | *Name* | N1 | N2 | N3 | N4 | N5 | N6 | N7 | N8 | N9 | N10 | N11 | N12 | - |
| | *Latitude* | 36.9 | 36.9 | 36.9 | 36.9 | 36.8 | 36.8 | 36.8 | 36.8 | 36.7 | 36.7 | 36.7 | 36.7 | - |
| | *Longitude* | 51.4 | 51.5 | 51.6 | 51.7 | 51.4 | 51.5 | 51.6 | 51.7 | 51.4 | 51.5 | 51.6 | 51.7 | - |
| Ramsar | *Point No.* | 61 | 62 | 63 | 64 | 65 | 66 | 67 | 68 | 69 | 70 | 71 | - | - |
| | *Name* | R1 | R2 | R3 | R4 | R5 | R6 | R7 | R8 | R9 | R10 | R11 | - | - |
| | *Latitude* | 37.2 | 37.2 | 37.2 | 37.2 | 37.1 | 37.1 | 37.1 | 37.1 | 37 | 37 | 37 | - | - |
| | *Longitude* | 50.5 | 50.6 | 50.7 | 50.8 | 50.5 | 50.6 | 50.7 | 50.8 | 50.6 | 50.7 | 50.8 | - | - |
| Kiashahr | *Point No.* | 72 | 73 | 74 | 75 | 76 | 77 | 78 | 79 | 80 | 81 | 82 | 83 | - |
| | *Name* | K1 | K2 | K3 | K4 | K5 | K6 | K7 | K8 | K9 | K10 | K11 | K12 | - |
| | *Latitude* | 37.7 | 37.7 | 37.7 | 37.7 | 37.6 | 37.6 | 37.6 | 37.6 | 37.5 | 37.5 | 37.5 | 37.5 | - |
| | *Longitude* | 49.8 | 49.9 | 50 | 50.1 | 49.8 | 49.9 | 50 | 50.1 | 49.8 | 49.9 | 50 | 50.1 | - |
| Anzali port | *Point No.* | 84 | 85 | 86 | 87 | 88 | 89 | 90 | 91 | 92 | 93 | 94 | - | - |
| | *Name* | Z1 | Z2 | Z3 | Z4 | Z5 | Z6 | Z7 | Z8 | Z9 | Z10 | Z11 | - | - |
| | *Latitude* | 37.7 | 37.7 | 37.7 | 37.7 | 37.6 | 37.6 | 37.6 | 37.6 | 37.5 | 37.5 | 37.5 | - | - |
| | *Longitude* | 49.4 | 49.5 | 49.6 | 49.7 | 49.4 | 49.5 | 49.6 | 49.7 | 49.4 | 49.5 | 49.6 | - | - |
| Astara port | *Point No.* | 95 | 96 | 97 | 98 | 99 | 100 | 101 | 102 | 103 | 104 | 105 | - | - |
| | *Name* | S1 | S2 | S3 | S4 | S5 | S6 | S7 | S8 | S9 | S10 | S11 | - | - |
| | *Latitude* | 38.4 | 38.4 | 38.4 | 38.3 | 38.3 | 38.3 | 38.2 | 38.2 | 38.2 | 38.1 | 38.1 | - | - |
| | *Longitude* | 48.9 | 49 | 49.1 | 48.9 | 49 | 49.1 | 48.9 | 49 | 49.1 | 49 | 49.1 | - | - |

The sea surface elevation is a random function of time. Its total variance is:

$$\overline{\xi^2} = \int_0^\infty S_f\,(f)df \qquad (4)$$

and the energy density spectrum is defined as [44]:

$$E_f(f) = \rho g S_f(f) = \frac{1}{2}\rho g \lim_{\Delta f \to 0} \frac{1}{\Delta f}\overline{a_i^2} \tag{5}$$

When the random sea-surface elevation is treated as a stationary, Gaussian process, then all statistical characteristics are determined by the variance density spectrum. These characteristics will be expressed in terms of the moments of the spectrum (moment of order m):

$$m_n = \int_0^\infty f^n\, S_f(f)df \quad (m = \cdots, -3, -2, -1, 0, 1, 2, 3, \dots) \tag{6}$$

Energy flux of irregular wave based on spectrum analysis and moments of the spectrum is defined as [44]:

$$d\bar{P}_{irr} = c_g E_f(f)df = \rho g c_g S_f(f)df = \frac{\rho g^2}{4\pi} S_f(f)\frac{1}{f}df \tag{7}$$

By integration [45]:

$$\bar{P}_{irr} = \frac{\rho g^2}{4\pi}\int_0^\infty S_f\,(f)\frac{1}{f}df = \frac{\rho g^2}{4\pi}m_{-1} \tag{8}$$

We will use this equation as the main analytical equation for determining the energy of the points in the examined region. Based on the spectrum analysis, the average values of wave power in each point will be obtained. (because of the depth of the examined points, the value of $n=C_g/C=0.5$ was assumed.)

### 3.3Step 3: Developing and evaluating the application of the GWO algorithm

In this approach, using the regular wave theory, the wave power equation was extracted and, by the GWO optimization algorithm, the optimal values of its parameters were specified. In this procedure, a point, the parameters of which had the closest values to the optimal parameters, was specified as the point with maximum wave energy. Therefore, first, by performing the time hystory analysis, values of depth and averaged wave height and averaged period over a time interval were obtained, and the following equation was taken into account for calculating the wave energy flux as the main equation in optimization algorithm [46]:

$$\bar{P}_{reg} = \bar{E}nC_0 \tanh(kd) \tag{9}$$

Where $E$ and $C_0$ are total averaged wave energy over a time interval and wave celerity in deep water condition. Therefore, in a general condition, equation 9 can be rewritten as:

$$\bar{P}_{reg} = \frac{\rho g^2 \bar{H}^2 \bar{T} \tanh(kd)}{32\,\pi}(1 + \frac{2kd}{\sinh(2k\ \ )}) \quad (10)$$

The wave number ($k$) is related to the wave period by dispersion equation [47].

The main goal of optimization was to calculate maximum values of $P_{reg,}$ and optimum values of $T$, $H$ and $d$ (i.e. $T_{opt}$, $H_{opt}$, $d_{opt}$). The optimization algorithm performed this search from among minimum and maximum values of $T$, $H$ and $d$. Thus, the boundary equations used in optimization are:

$$\bar{H}_{min} < \bar{H}_{opt} < \bar{H}_{max}\ ,\ \bar{T}_{min} < \bar{T}_{opt} < \bar{T}_{max}\ ,\ d_{min} < d_{opt} < d_{max} \quad (11)$$

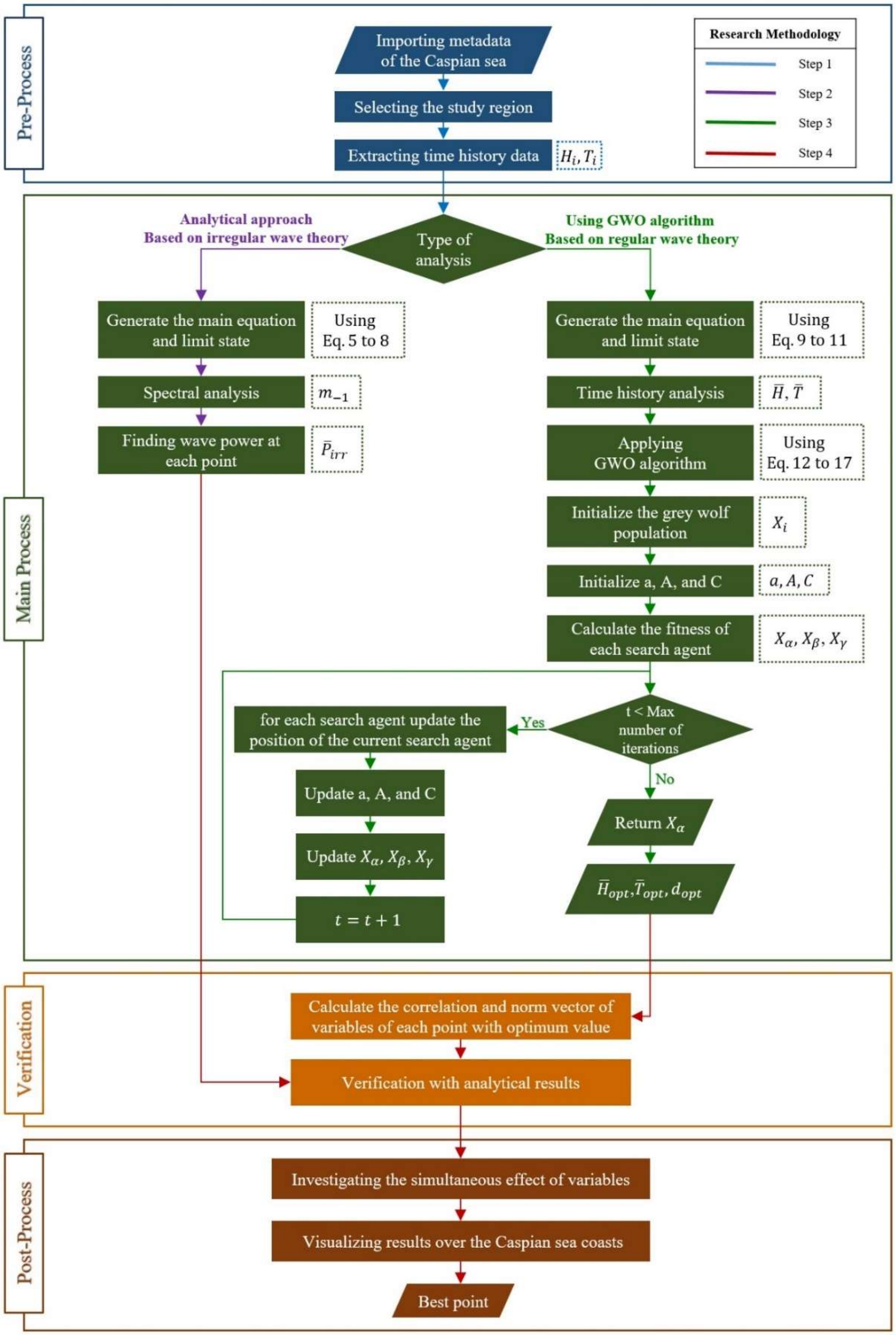

**Figure 6:** The flowchart of the proposed optimization framework

For searching in this range, the grey wolf optimizer (GWO) was employed. In this algorithm by performing a simulation, searching for the prey, encircling the prey, and mass attacking the prey are presented in the form of a mathematical model. For mathematical modeling of the wolves' prey behavior, while developing the GWO, the best solution is called wolf $\alpha$. Thus, the second and third best solutions are, respectively named wolf $\beta$ and $\delta$. It is assumed that the remaining solutions are $\omega$. Thus, in the GWO algorithm, optimization is led by $\alpha$, $\beta$ and $\delta$, and the $\omega$ wolves follow these three groups and improve their distance from the other three groups using vector $\vec{D}$. As noted above, grey wolves encircle the prey while searching. To mathematically model this encircling behavior, the following equations are proposed [32]:

$$\begin{aligned} \vec{D}_\alpha &= \left|\vec{C}_1\, \vec{X}_\alpha - \vec{X}\right| \\ \vec{D}_\beta &= \left|\vec{C}_2\, \vec{X}_\beta - \vec{X}\right| \\ \vec{D}_\delta &= \left|\vec{C}_3\, \vec{X}_\delta - \vec{X}\right| \end{aligned} \tag{12}$$

$$\begin{aligned} \vec{X}_1 &= \vec{X}_\alpha - \vec{A}_1(\vec{D}_\alpha) \\ \vec{X}_2 &= \vec{X}_\beta - \vec{A}_2(\vec{D}_\beta) \\ \vec{X}_3 &= \vec{X}_\delta - \vec{A}_3(\vec{D}_\gamma) \end{aligned} \tag{13}$$

$$\vec{X}_{(t+1)} = \frac{\vec{X}_1+\vec{X}_2+\vec{X}_3}{3} \tag{14}$$

Where $t$ is the current iteration, $\vec{A}$ and $\vec{C}$ are the vectors of coefficients, $X_p$ is the prey's position vector, and $X_i$ is a grey wolf's position vector. Vectors $\vec{A}$ and $\vec{C}$ are calculated with the following equations:

$$\vec{A} = 2\vec{a}.\vec{r}_1 - \vec{a} \tag{15}$$

$$a = 2 - iter.\left(\frac{2}{Max_{iter}}\right) \tag{16}$$

$$\vec{C} = 2\,\vec{r}_2 \tag{17}$$

Where $a$ component is reduced linearly by increasing the number of iterations (from 2 to 0) and $r_1$ and $r_2$ are random vectors (between 0 and 1). By continuing this trend repeatedly in the examined range, vector $A$ is reduced per $a$. In other words, $A$ is a random value within (-2, 2) range, where $a$ is reduced from 2 to 0 by increasing the number of iterations. When the random values of $A$ are in the range (-1, 1), the next position of a search factor can be any position between the current position and that of the prey. Finally, achieving the in equation $|A| < 1$ force the wolves to attack the prey, the algorithm is ended, and the final solutions are introduced as optimized parameters[32]. This algorithm has been used as a method for finding the best optimized values of $T$, $H$ and $d$, which maximize $P_{reg.}$ These values for $T_{opt}$, $H_{opt}$, $d_{opt}$ will be taken as a criterion to predict the power potency of waves in each point. Therefore, the norm of subtraction vector and the correlation of each point's properties with optimum values will be calculated. Thus, the point with the greatest correlation and lowest norm value will be specified as the best point. The Euclidean norm of $x \in \mathbb{R}^n$ is defined as:

$$\|\boldsymbol{X}\|_2 = \sqrt{\sum_{i=1}^{n} |\boldsymbol{X}|_i^2} \tag{18}$$

Where in this problem vector $|\boldsymbol{X}|_i$ is:

$$|\boldsymbol{X}|_i = |\boldsymbol{averaged}|_i - |\boldsymbol{opt}|_i \tag{19}$$

$$|\boldsymbol{averaged}|_i = |\overline{H}_i, \overline{T}_i, d_i| \,, |\boldsymbol{opt}|_i = |H_{opt}, T_{opt}, d_{opt}| \tag{20}$$

**3.4 Step 4: Forming the numerical model and comparing the results**

After finding the average power of waves at all 105 points, the results of the analytical method (based on irregular wave theory) and the solution method using the optimization algorithm (based on regular wave theory) are compared and the best point in the southern coasts of the Caspian Sea with maximum wave energy is introduced. The output of these modeling can help find an appropriate point for installing the wave energy converter. Figure 6 can present the flowchart of this study.

# 4 Result and discussion

According to the research method mentioned in the previous section, the results will be presented in two parts. First, the wave power outputs at all points will be given and compared using analytical equations based on irregular wave theory. This comparison will also be done for all points within each zone. In the next section, using the GWO optimization algorithm, the optimal values of $H_{s_{opt}}$, $T_{e_{opt}}$ and $d_{opt}$ are found, such that the maximum power available based on regular wave theory and use of time-history analysis would be obtained.

**4.1 Analitycal results**

Initially, the wave power along the southern coast of the Caspian Sea is calculated. An overview of the power of the waves for comparison is shown in Fig.7. As shown in the figure, the power of the waves along the southwest coast is greater than the power of the waves on the southeast coast. This relative increase is larger, even up to 3 times. For example, on the one hand, we almost see that all Kiashahr points have high wave power, while on the other hand, they have slight power at all points in the Torkaman port. Therefore, the southeast coast is not a suitable place for establishing the wave energy converter system. However, the southwestern coasts (especially Anzali port and Kiashahr coasts) seem to be good places to use the waves energy. It seems logical by comparing Fig. 7. with Fig. 2 and 3, as the points around the western ports, are also deeper and the relative increase in the height of the waves in the area can result in increasing the available energy in the area. The waves also generate different power at each point. For better comparison, the values of the wave power at points in each range are shown in Fig. 8.

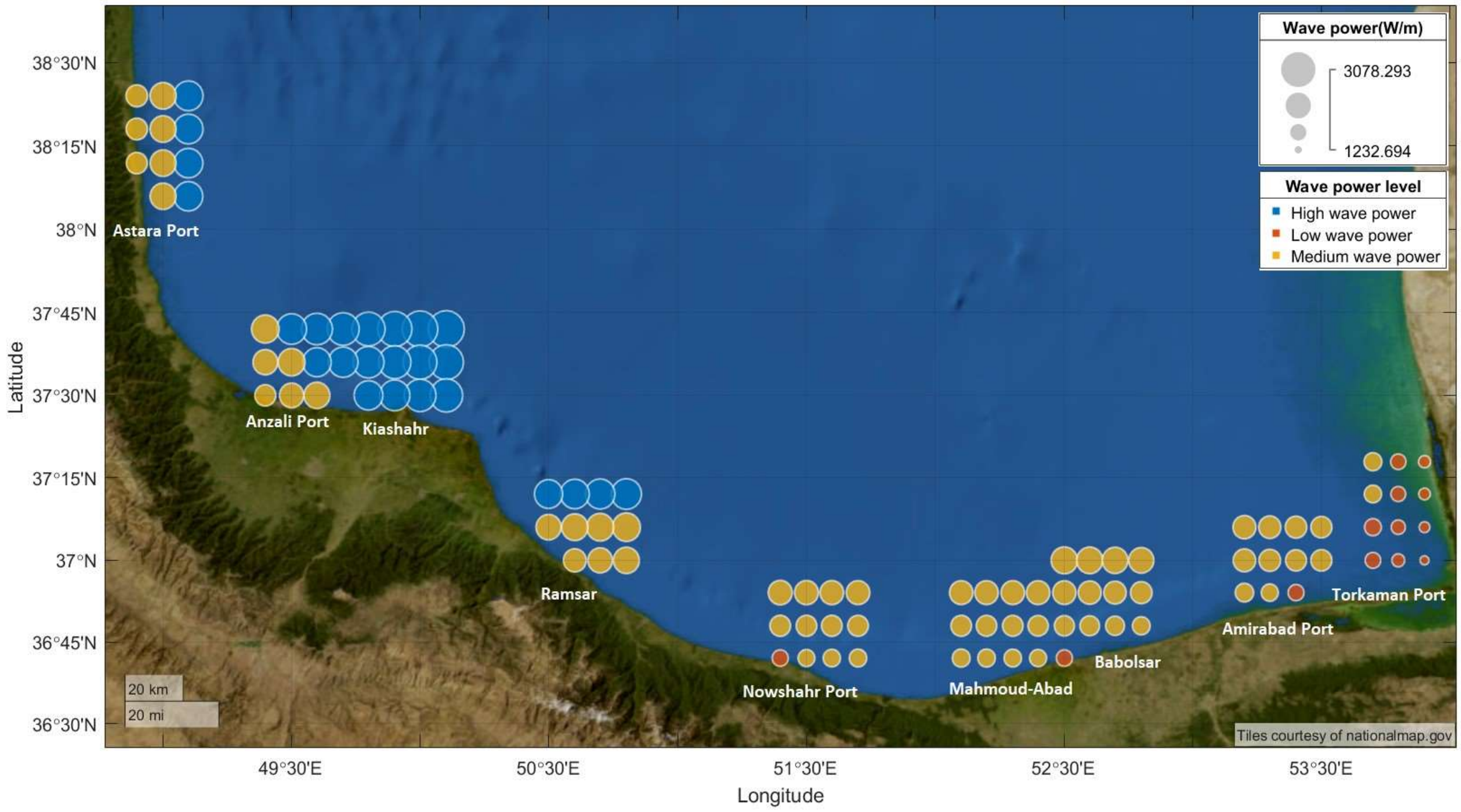

**Figure 7:** The wave power of all 105 examined points in the South of the Caspian Sea.

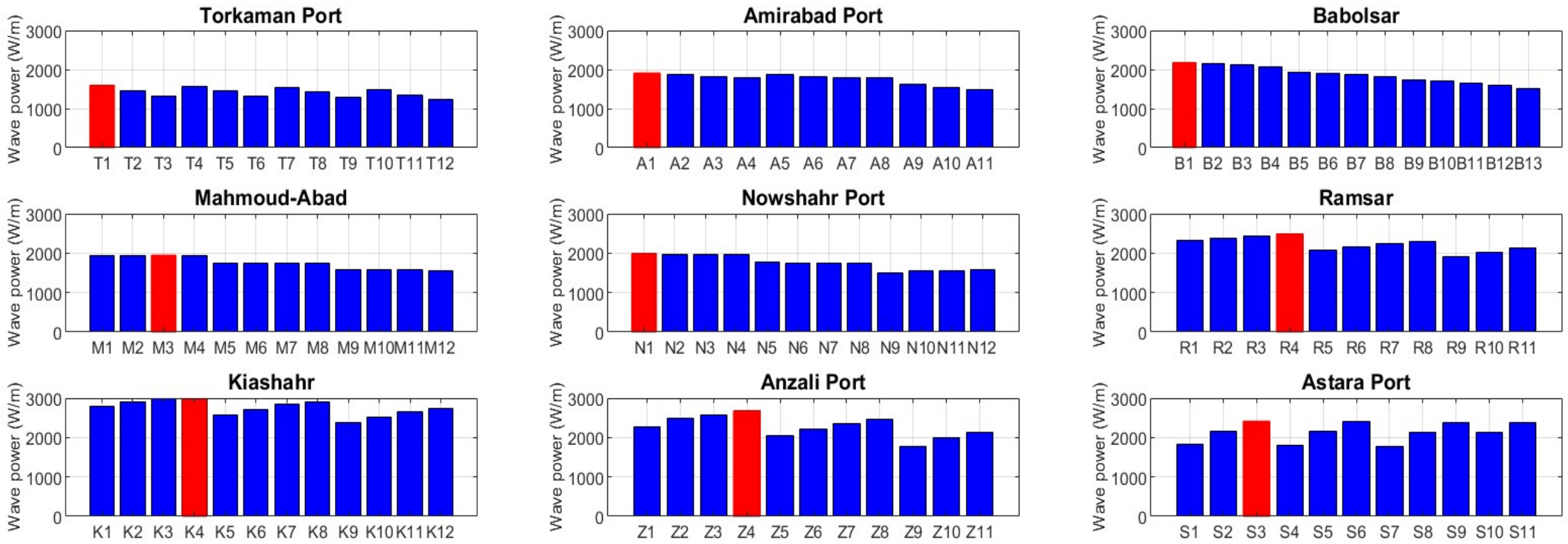

**Figure 8:** The comparison of wave power among all 105 examined points in the South of the Caspian Sea.

After examining and comparing the energy of the points within each range, it can be said that in Torkaman Port, the range of point $T_1$ with the wave power of 1594 W/m has the maximum wave power among the other ranges. Similarly, in Amirabad Port, point A1 with 1914 W/m, in Babolsar area point $B_1$ with 2181 W/m, in Mahmoud-Abad area point $M_3$ with 1952 W/m, in Nowshahr area point $N_1$ with 1988W/m, in Ramsar area point $R_4$ with 2489 W/m, in Kiashahr area point $K_4$ with 3074 W/m, in Anzali Port area point with 2676 W/m, and point $S_3$ in Astara Port with 2414 W/m have the most wave power. These points are marked in red on the Fig. 8 As a result; it can be seen that the offshore points have greater energy potential than the onshore points. Fig. 9 shows the wave energy variations at each given $H_s$ and $T_e$.

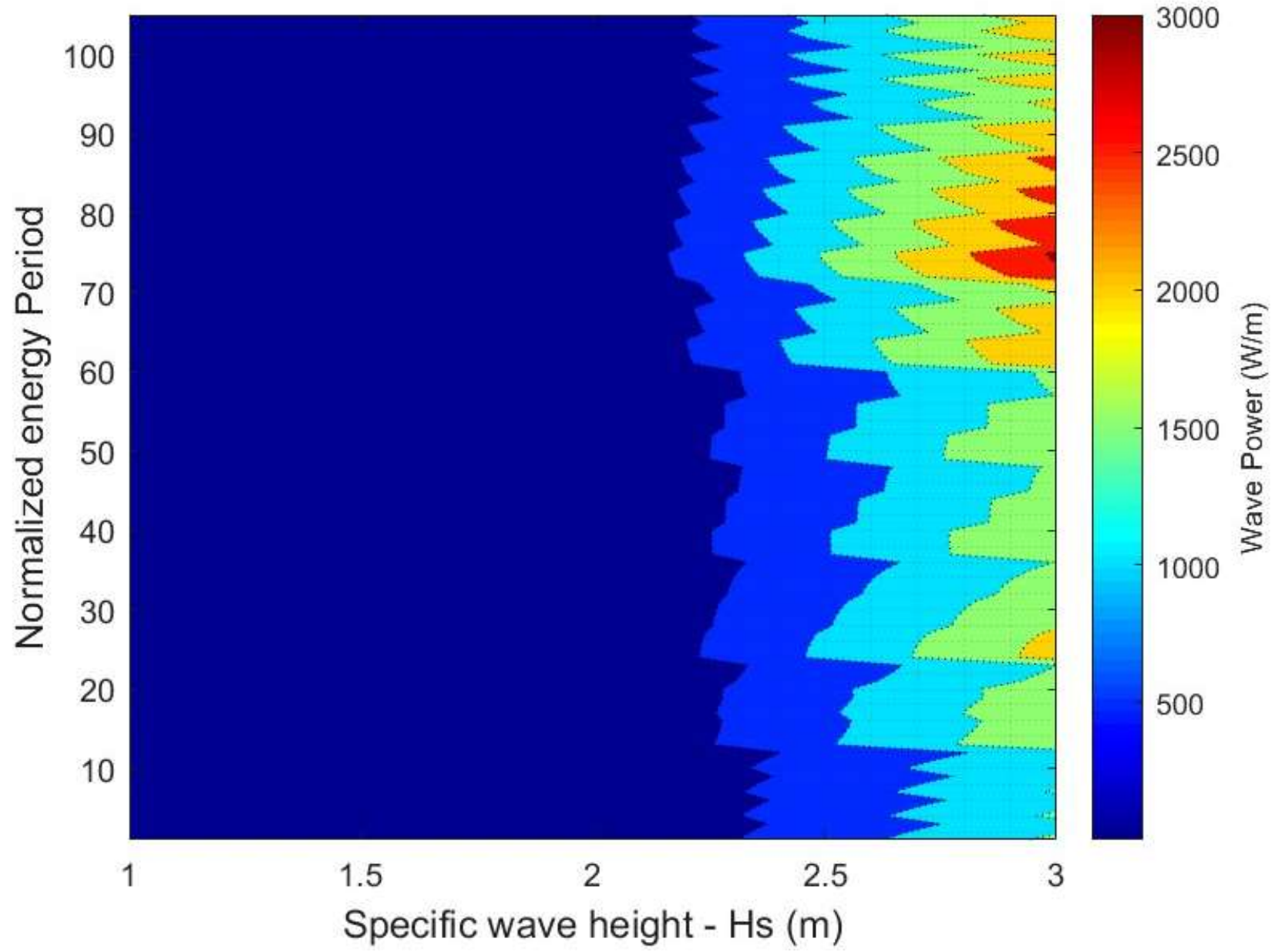

**Figure 9:** The distribution of the wave power in the South of the Caspian Sea concerning wave height and wave period.

### 4.2. Optimisation results

By running the GWO optimization algorithm on different ranges of Equation 11, it began searching the optimal solutions for H, T, and d, which maximize the energy in limitations of Equation 10. The number of search agents and maximum iteration for finding the solution was, respectively, 10 and 200. It was observed that the best solution from searching the alpha agent was stabilized after 80 iterations. Fig. 10 shows the convergence curve of results.

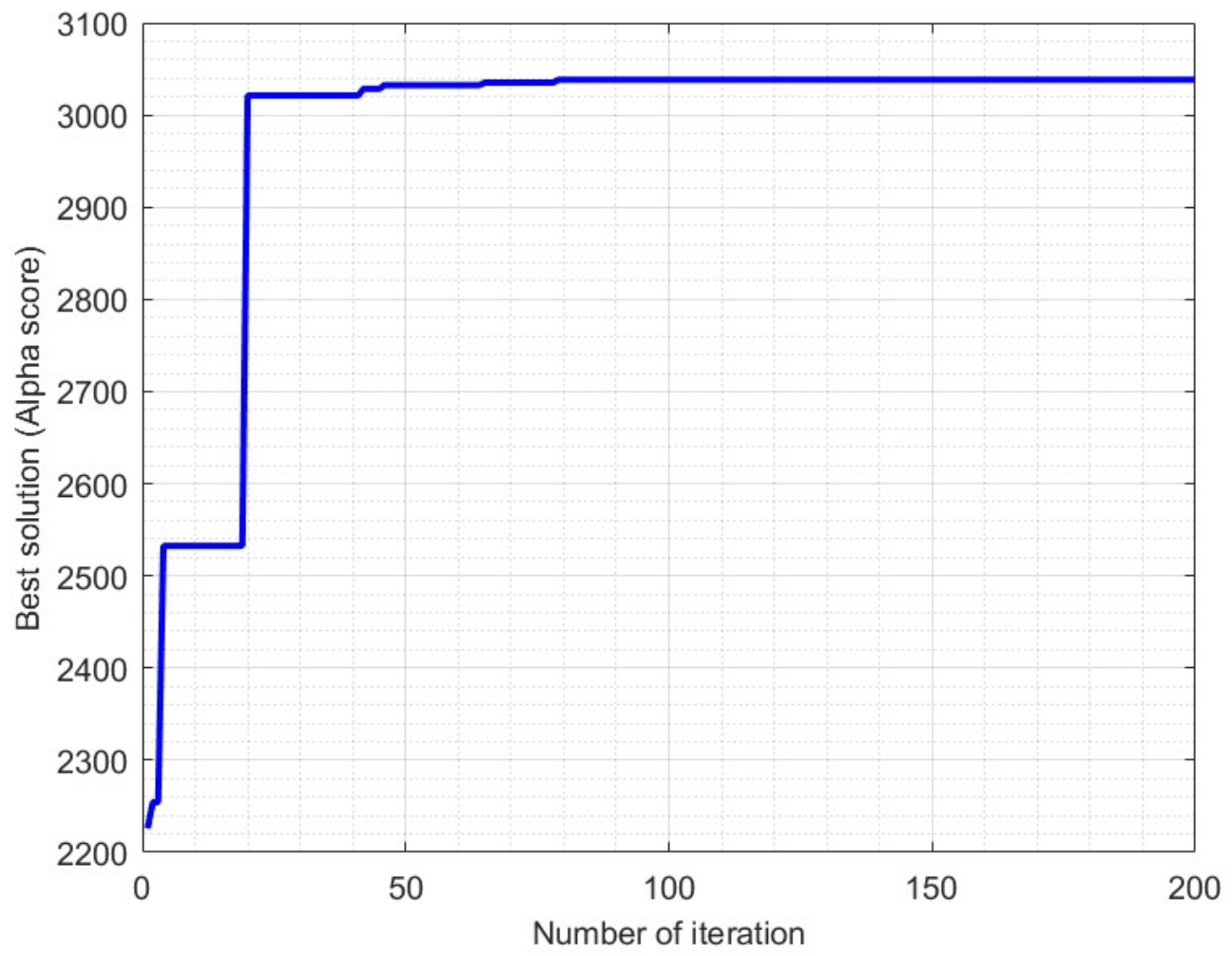

**Figure 10:** convergence curve of GWO optimization algorithm for searching in limitations of equation 11.

It was observed that the solution stability in iterations between 3-18 and 20-45 was due to getting stuck in local optimal solutions. As a result, the number of 200 iterations seems adequate. The optimal solutions within the southern coasts of the Caspian Sea were obtained using the optimization algorithm and shown in Table 2.

**Table 2:** The optimum values of the height, depth and period of the wave.

| *Variable* | *Optimum Value* | *Unit* |
|---|---|---|
| $\bar{H}_{opt}$ | 0.595 | *(m)* |
| $\bar{T}_{opt}$ | 4.102 | *(s)* |
| *depth* | 79.218 | *(m)* |

These values can be used to investigate the competence of different points in determining the highest wave energy. Based on this, the norm value of the difference between the parameters of each point and provided optimal values were investigated. The lower norm value showed the highest level of wave energy on that point. Finally, the point with the lowest norm value was selected as the best point for implementing the wave energy transformation system. Fig. 11 shows the norm vector size based on the Equations 17-19 for the points under investigation. It shows $4^{th}$ point in “Kiashahr” (K4) as the best point, which is consistent with the analytical results from Fig.8 For general verification of the proposed approach, the correlation coefficients of different points were compared to the wave power values (output of the irregular wave model) in Fig. 12. The closer is the correlation values to one, the greater similarity is observed between the characteristics of the point under investigation and those of the optimal point. That shows a good match between the correlation of the points (output of the GWO optimization algorithm) and wave power values (output of the analytical approach) in terms of general variations. A greater correlation indicates a greater amount of energy on each point; whereas, a lower correlation indicates a lower amount of energy on the point under investigation. This finding suggests that the developed model can be used as a solution to assess these points and determine the fitness of them in achieving maximum energy. As was mentioned, this method used the time history analysis, which has a lower level of complexity and computational time, to find the best point for the implementation of the wave energy converter system.

For investigate the wave energy variations with changes in the significant wave height and depth, these three parameters were presented together in Fig.13.

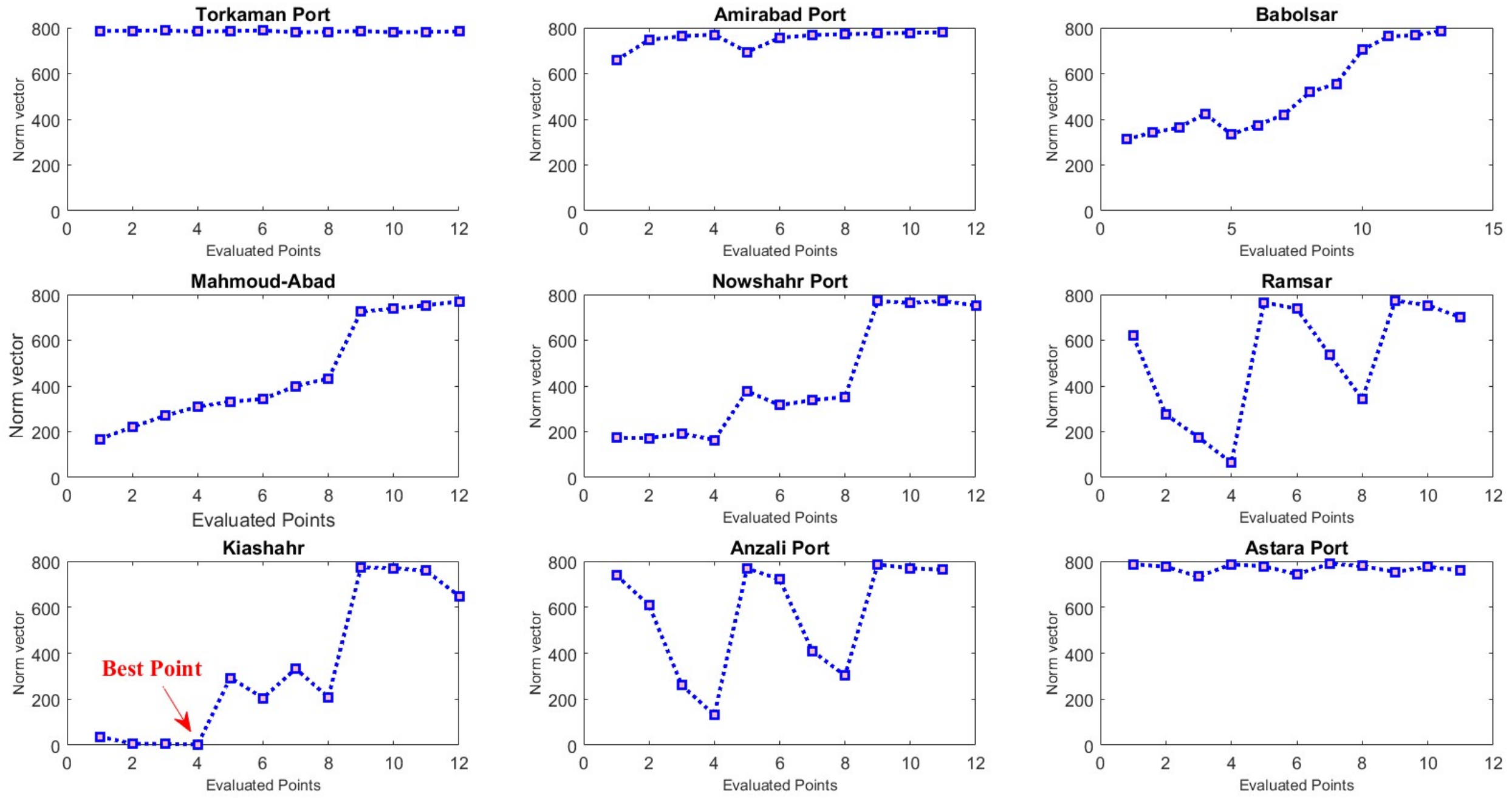

**Figure 11:** Norm vector values of examined locations' outputs with the optimum solution at each zone.

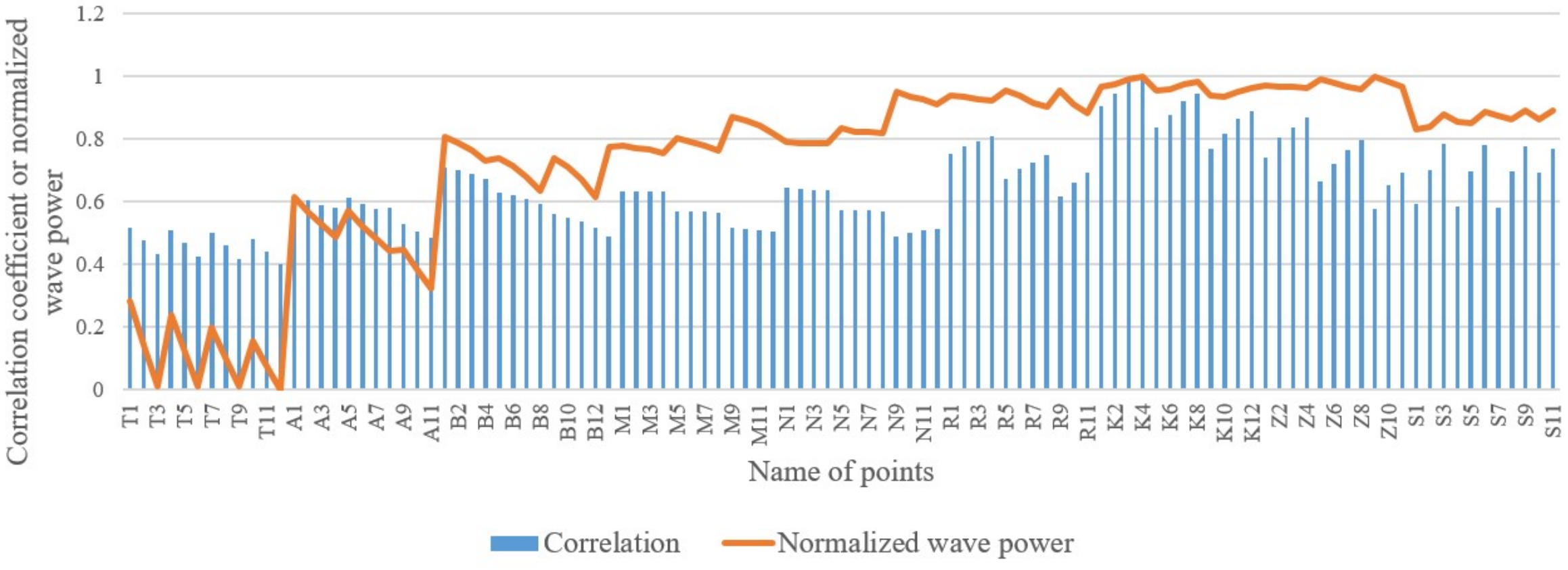

**Figure 12:** Comparison of correlation coefficient with normalized wave power for each evaluated point.

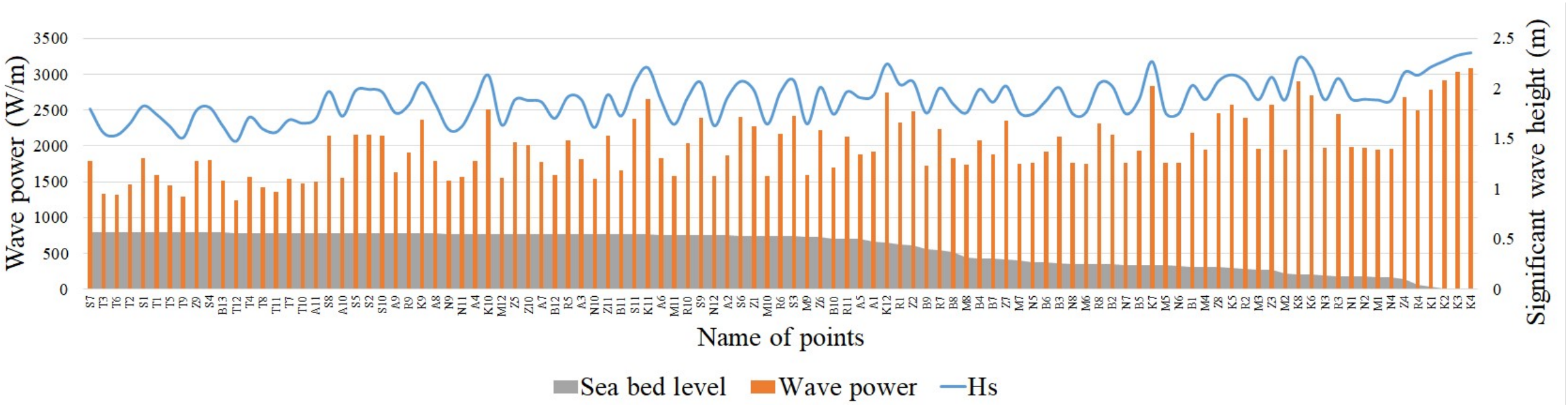

**Figure 13:** Trend of wave power and significant wave height over the change of depth.

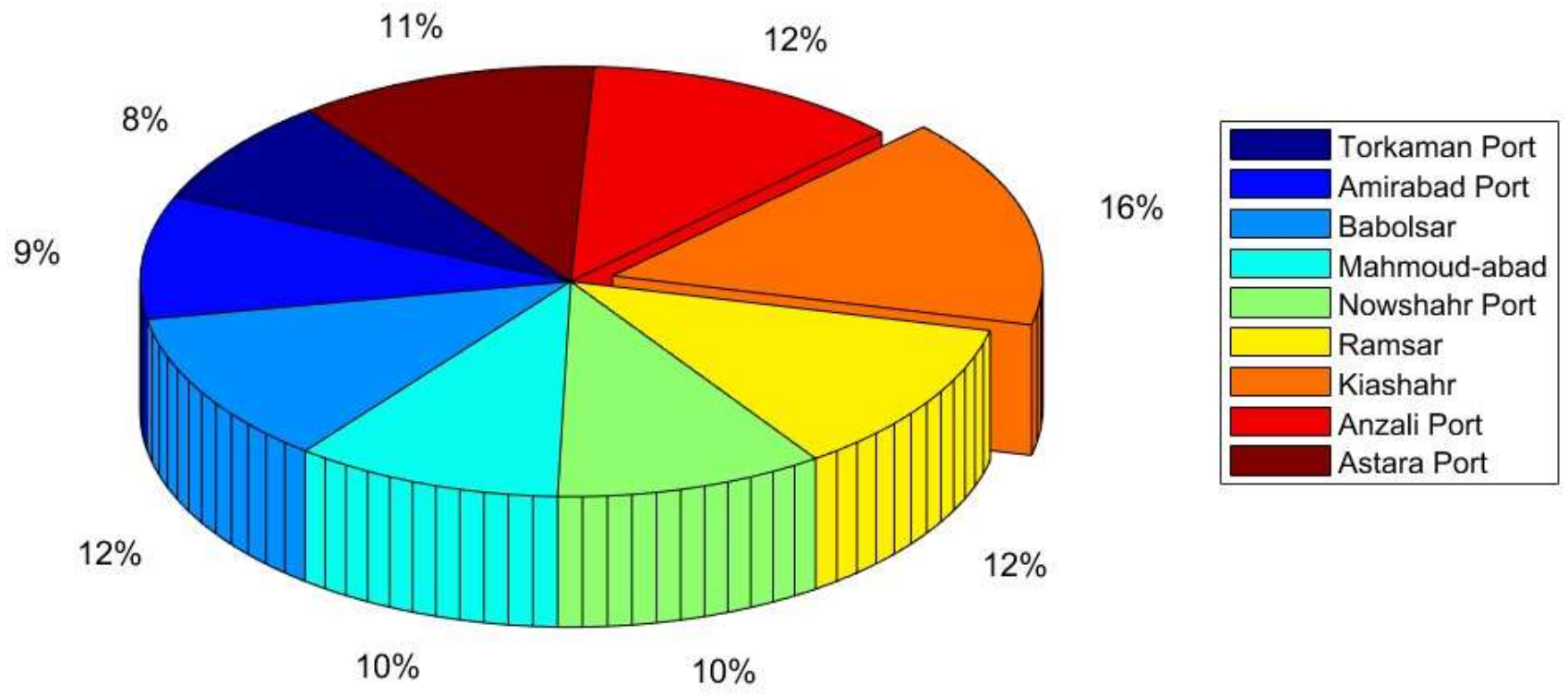

**Figure 14:** comparison of each port's wave power portion of total energy.

Investigating the depth of the points to the wave power on each point, as a predictable trend of change, showed that the wave power is typically higher on the deeper points. In other words, the onshore points have a lower level of wave power. It is worth noting that although the amount of offshore energy is greater than the onshore energy, the implementation of the wave energy converter systems is costly because of accessibility. Moreover, investigating the concurrent effect of the significant wave height and depth revealed that changes in the wave energy level are more influenced by the wave height than the wave depth. This finding can be justified by nonlinear dependence of the wave power on the wave height on each point. Finally, Fig. 14 shows each port's wave power portion of the total energy of southern investigated area of the Caspian Sea. According to this figure, the points with the highest amount of energy are located in Kiashahr Port. Also, due to the Fig. 8, Among 105 points selected along the southern coastline of the Caspian Sea, K4, K3, K2, K8, and K7 ports were respectively prioritized. The selected points can similarly be prioritized based on the comparative graph in Fig. 8 and Fig. 14.

# 5 Conclusions

Regarding the high potential of the Caspian Sea, notably higher wave power in its southern regions, in supplying the required energy of its coastal areas, comprehensive studies were conducted on the wave power potential in the south of Caspian. Some areas around nine ports along these coasts were selected, and their data was extracted. Analytical analysis based on irregular wave theory showed that the southwestern regions had a higher level of wave power than the southeastern areas. Moreover, the points within the Kiashahr and Anzali Ports had higher energy than the southern and southeastern regions. A new approach is developed to compare these points and measure their fitness in supplying the maximum wave energy using the GWO and time history analysis. In this way, the position with maximum wave energy can be predicted without calculating the wave power in each point based on the spectral analysis. In this method, the optimal parameters were extracted from the algorithm for the assessment of the locations within the southern areas of the Caspian Sea. These parameters are, respectively, $H_{opt} = 0.595(m)$, $T_{opt} = 4.102(s)$, and $d_{opt} = 79.218(m)$. These values were regarded as the assessment indices within the south of regions of the Caspian Sea. Then, the fitness of each point is obtained using the correlation function and the Euclidean norm vector to present the most optimal position with maximum potential for wave energy exploitation. Further, validation of the algorithm output with the output power using the analytical method suggested good capability of the new algorithm in predicting and comparing the wave power. The concurrent comparison of the parameters affecting the wave power showed that the wave power concentration occurs at $H_s = 3$ and $0.75 \times T_{e_{max}} < T_e < 0.85 \times T_{e_{max}}$. Moreover, the wave power in the southern coasts of the Caspian Sea was higher in deeper areas. In other words, the onshore points have lower wave power. Although the amount of offshore energy is greater than the onshore energy, the implementation of the wave energy converter systems is costly because of difficult accessibility. Investigating the concurrent effect of the significant wave height and depth revealed that changes in the wave energy level are influenced more by the wave height than the wave depth. It can be concluded that this method and its outputs can be used to compare the fitness of the points in supplying energy at each port with a certain depth and time history without the need for spectral analysis.

## Conflicts of Interest

The authors declare that there is no conflict of interest regarding the publication of this paper.

## Funding Statement

This work was supported and granted by Iran Ports & Maritime Organization (IPMO) with the grant number: TE178.

## Acknowledgement

The authors would like to express their gratitude from the Iranian National Institute for Oceanography and atmospheric science (INIO), which provided the data needed for performing this project. Also, we shall appreciate the support of Dr Mehdi Neshat, who was of great help, valuable comments and useful suggestions in this research.